\RequirePackage{fix-cm}
\documentclass[smallextended]{svjour3}       % onecolumn (second format)
\smartqed  % flush right qed marks, e.g. at end of proof
\usepackage{graphicx}
\usepackage{multirow}
\usepackage{subfigure}
\usepackage{colortbl,xcolor}
\definecolor{Gray}{gray}{0.8}
\usepackage{amsmath,amssymb} % define this before the line numbering.
\usepackage{color}
\def\W{\mathbf{W}}
\def\A{\mathbf{A}}

\def\ie#1{\textit{i.e.}{#1}}
\def\eg#1{\textit{e.g.}{#1}}
\long\def\comment#1{}
\usepackage{xspace}

\begin{document}

\title{Bi-Real Net: Binarizing Deep Network towards Real-Network Performance}
%\subtitle{Binarizing Deep Network Towards Real-Network Performance}
%\titlerunning{Short form of title}        % if too long for running head

\author{Zechun Liu \and
Wenhan Luo \and
Baoyuan Wu \and
Xin Yang \and
Wei Liu \and
Kwang-Ting Cheng}

%\authorrunning{Short form of author list} % if too long for running head

\institute{Zechun Liu \at
              Hong Kong University of Science and Technology, Hong Kong, China \\
              \email{liuzechun0216@gmail.com} 
           \and
           Wenhan Luo \at
              Tencent AI lab, Shenzhen, China\\
              \email{ whluo.china@gmail.com}
           \and
           Baoyuan Wu \at
              Tencent AI lab, Shenzhen, China\\
              \email{wubaoyuan1987@gmail.com}
           \and
           Xin Yang \at
              Huazhong University of Science and Technology, Wuhan, China\\
              \email{xinyang2014@hust.edu.cn}
           \and
           Wei Liu \at
              Tencent AI lab, Shenzhen, China\\
              \email{wl2223@columbia.edu}
           \and
           Kwang-Ting Cheng \at
              Hong Kong University of Science and Technology, Hong Kong, China \\
              \email{timcheng@ust.hk}
}

\date{Received: date / Accepted: date}
% The correct dates will be entered by the editor

\maketitle

\begin{abstract}

In this paper, we study 1-bit convolutional neural networks (CNNs), of which both the weights and activations are binary. While being efficient, the lacking of a representational capability and the training difficulty impede 1-bit CNNs from performing as well as real-valued networks. To this end, we propose Bi-Real net with a novel training algorithm to tackle these two challenges. To enhance the representational capability, we propagate the real-valued activations generated by each 1-bit convolution via a parameter-free shortcut. To address the training difficulty, we propose a training algorithm using a tighter approximation to the derivative of the sign function, a magnitude-aware binarization for weight updating, a better initialization method, and a two-step scheme for training a deep network. Experiments on ImageNet show that an 18-layer Bi-Real net with the proposed training algorithm achieves 56.4\% top-1 classification accuracy, which is 10\% higher than the state-of-the-arts ({\it e.g.}, XNOR-Net), with a greater memory saving and a lower computational cost. Bi-Real net is also the first to scale up 1-bit CNNs to an ultra-deep network with 152 layers, and achieves 64.5\% top-1 accuracy on ImageNet. A 50-layer Bi-Real net shows comparable performance to a real-valued network on the depth estimation task with merely a 0.3\% accuracy gap.
% \PACS{PACS code1 \and PACS code2 \and more}
% \subclass{MSC code1 \and MSC code2 \and more}
\end{abstract}

\section{Introduction}
Deep Convolutional Neural Networks (CNNs) have achieved substantial advances in a wide range of vision tasks, such as object detection and recognition~\cite{alexnet,vggnet,googlenet,inception,resnet,rcnn,faster-rcnn}, depth perception~\cite{unsupervised-cnn-depth,monocular-depth}, visual relation detection~\cite{zhang2017relation-1,zhang2017relation-2}, face tracking and alignment~\cite{facial-detection1,facial-localization2,facial-alignment,wu-face-tracking-iccv,wu-face-tracking-pr}, object tracking~\cite{luo2018end}, etc.
However, the superior performance of CNNs usually requires powerful hardware with abundant computing and memory resources, for example, high-end graphics processing units (GPUs). 
Meanwhile, there are growing demands to run vision tasks, such as augmented reality and intelligent navigation, on mobile hand-held devices and small drones. The CNN models are usually trained on GPUs and deployed on mobile devices for inference. Most mobile devices are equipped with neither powerful GPUs nor adequate memory to store and run expensive CNN models. 
Consequently, the high demand for computation and memory becomes the bottleneck of deploying powerful CNNs on most mobile devices. 
In general, there are two major approaches to alleviate this limitation. The first is to reduce the number of weights with more compact network design or pruning. The second is to quantize the weights or quantize both the weights and activations, with the extreme case of both the weights and activations being binary. 

In this work, we study the extreme case of the second approach, {\it i.e.}, one binary CNN. 
It is also called 1-bit CNN, as each weight parameter and activation can be represented by a single bit. 
As demonstrated in~\cite{xnornet},  up to a $32 \times$ memory saving and a $58 \times$ speedup on CPUs have been achieved for a 1-bit convolutional layer, in which the computationally heavy matrix multiplication operations can be implemented using light-weighted bitwise XNOR operations and popcount operations. The current binarization methods achieve comparable accuracy to real-valued networks on small datasets ({\it e.g.}, CIFAR-10 and MNIST). However, on large-scale datasets ({\it e.g.}, ImageNet), the binarization method based on AlexNet in \cite{binarynet} encounters a severe accuracy degradation, from $56.6\%$ to $27.9\%$ \cite{xnornet}.
This suggests that the capability of conventional 1-bit CNNs is not sufficient to cover the great diversity in large-scale datasets like ImageNet. Another binary network called XNOR-Net \cite{xnornet} enhances the performance of 1-bit CNNs by utilizing the absolute mean of weights and activations. XNOR-Net improves the accuracy to 44.2\% on AlexNet, which is encouraging, but there still remains a performance gap regarding the real-valued networks.

\begin{figure}[t]
\centering
\includegraphics[width=1\linewidth]{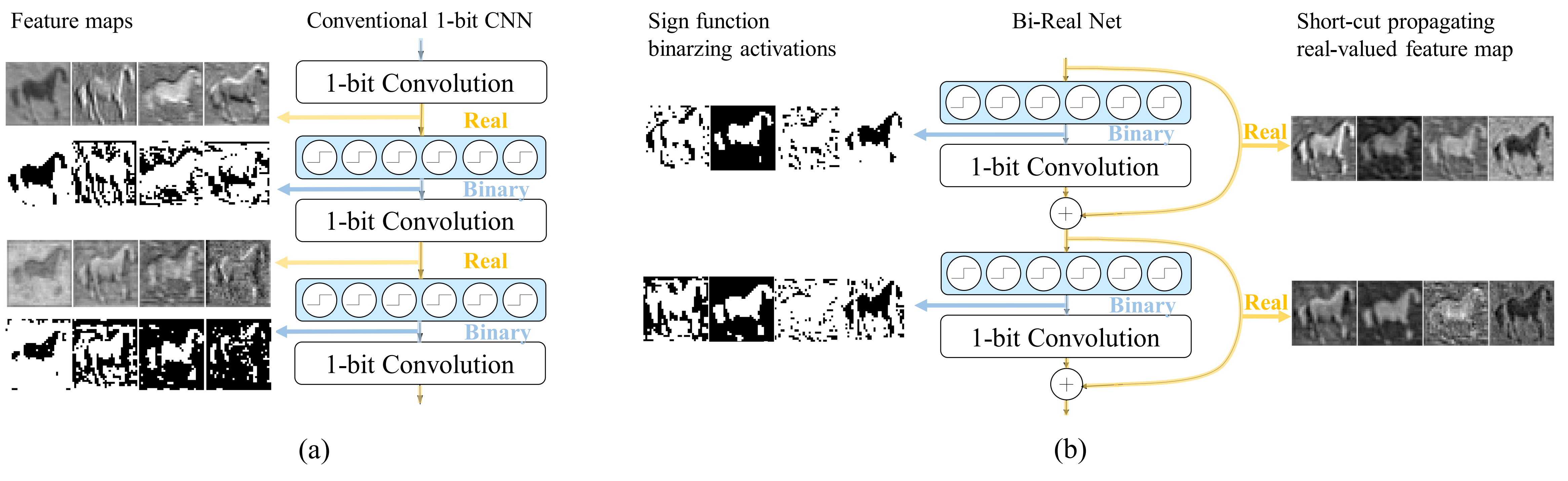}
\caption{Networks with intermediate feature visualization. Yellow lines denote real values propagated inside the path, while blue lines denote binary values. (a) A 1-bit CNN without shortcuts, and (b) the proposed Bi-Real net with shortcuts propagating real-valued features. }
\label{fig:shortcut_or_not}
\end{figure}

The objective of this study is to further improve 1-bit CNNs, as we believe that their potential has not yet been fully explored. 
One important observation is that during the inference process, a 1-bit convolutional layer generates integer outputs, as a result of the popcount operations. The integer outputs become real values after a BatchNorm \cite{batchnorm} layer. These real-valued activations are then binarized to $-1$ or $+1$ through the consecutive sign function, as shown in Fig. \ref{fig:shortcut_or_not}(a).
Obviously, compared to binary activations, these integers or real activations contain more information, which is lost in conventional 1-bit CNNs \cite{binarynet}.
Based on this observation, we propose to use parameter-free shortcut paths to collect all the real-valued outputs from each 1-bit convolutional layer, and add them to the next most adjacent real-valued outputs with matched dimensions. This proposed network is dubbed Bi-Real net, with real-valued shortcut paths transmitting high-precision features and efficient 1-bit convolutional layers extracting new features, respectively. Because ResNet \cite{resnet} is the most prevalent network structure, we verify our shortcut design principle on the shallow ResNet-based structures as well as the deep ResNet with the bottleneck structure.

For a shallow ResNet-based structure, we propose to add shortcuts forwarding the real activations to be added with the real-valued activations of the next block, as shown in Fig.  \ref{fig:shortcut_or_not}(b). By doing so, the representational capability of the proposed model becomes much greater than that of the original 1-bit CNNs, with only a negligible computational cost incurred by the extra element-wise addition and without any additional memory cost. This shortcut design results in a so-called 1-(convolutional-)layer-per-block structure, which is more effective than the 2-layer-per-block structure proposed in ResNet. The original ResNet argues that a shortcut has to bypass at least two convolutional layers; however, we provide a mathematical explanation of the feasibility of the 1-layer-per-block structure in Sec. \ref{sec:1-layer-per-block-explanation}

For a deep ResNet-based structure with the bottleneck, we propose to add another shortcut path in addition to the original shortcut path in ResNet. This shortcut path adds the input activations to the sign function before the 3$\times$3 convolutional layer and the output activations of the 3$\times$3 convolutional layers in series, as shown in Fig. \ref{fig:deeper_network}. This additional shortcut works jointly with the original shortcut collecting all the real-valued outputs in the 1-bit CNN, and the representational capability is thus greatly enhanced.

We further propose a novel training algorithm for 1-bit CNNs including four special technical features:

\begin{figure}[t]
\centering
\includegraphics[width=1\linewidth]{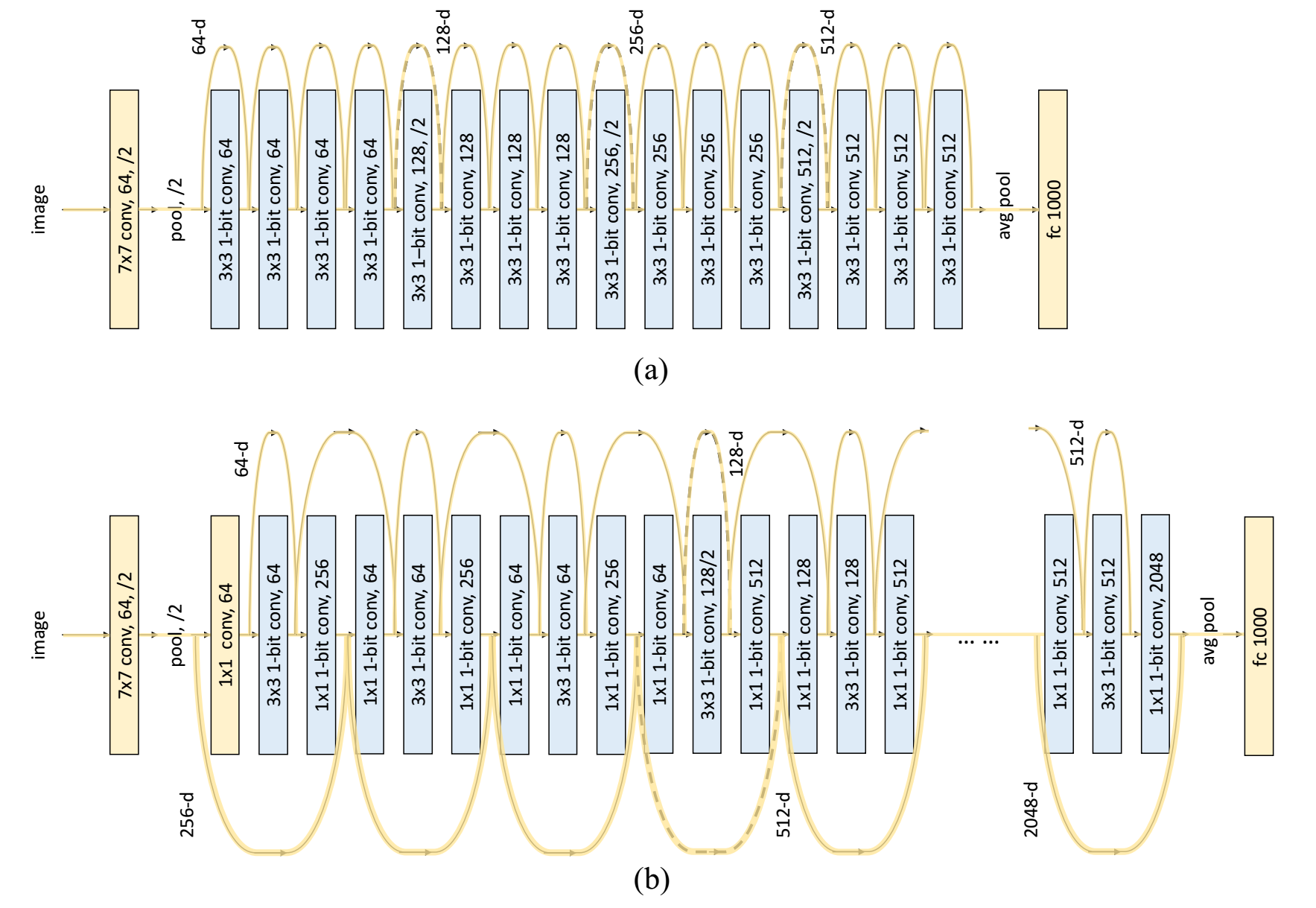}
\caption{The proposed network structure. (a) The shallow Bi-Real net for 18-layer and 34-layer structures, and (b) the deep Bi-Real net for 50-layer and 152-layer structures. The dashed lines denote a 2x2 average pooling layer followed by a 1$\times$1 convolutional layer for down-sampling and dimension matching.}
\label{fig:deeper_network}
\end{figure}

\begin{itemize}

\item \textbf{Magnitude-aware binarization with respect to weights.} As the gradient of the loss with respect to the binary weight will not be large enough to trigger the change to the sign of the binary weight, the binary weight cannot be directly updated using the standard gradient descent algorithm. In BinaryNet \cite{binarynet}, the real-valued weight is first updated using gradient descent, and the new binary weight is then obtained through taking the sign of the updated real weight. However, we observed that the gradient with respect to the real weight only depends on the sign of the current real weight, while independent of its magnitude. To derive a more effective gradient, we propose to use a magnitude-aware sign function during training, resulting in the desired dependence of the gradient with respect to the real weight on both the sign and the magnitude of the current real weight. After convergence, the binary weight (\ie, -1 or +1) is obtained through the sign function of the final real weight for inference. 

\item \textbf{Initialization.} As a highly non-convex optimization problem, training 1-bit CNNs could be sensitive to initialization. In \cite{dji}, the 1-bit CNN model is initialized using the real-valued CNN model with the ReLU function pre-trained on ImageNet. We propose to replace ReLU by the clip function in pre-training, as the activation of the clip function is closer to the binary activation than that of ReLU. 

\item \textbf{A two-step training method for deep 1-bit CNNs with the bottleneck structure.} As the network goes deeper, training becomes more difficult. Following the practice in multi-step training in quantizing the network~\cite{zhuangbohan}, we customize a two-step training method for our deep Bi-Real net with the bottleneck structure to ease the training difficulty. We first binarize the weights and activations in the 1$\times$1 convolutional layers and the activations in the 3$\times$3 convolutional layers. After the network converges, we further binarize the weights in the 3$\times$3 convolutional layers. This training procedure uses the real-valued weights in the 3$\times$3 convolutional layers as a transition for training the deep Bi-Real net, helping the network converge to reach higher accuracy.

\end{itemize}

Experiments on ImageNet show that these ideas are useful to train 1-bit CNNs. 
With the dedicatedly-designed shortcut and the proposed optimization techniques, our Bi-Real net, with only binary weights and activations inside each 1-bit convolutional layer, achieves 56.4\% and 62.2\% top-1 accuracy on the ImageNet dataset with 18-layer and 34-layer structures, respectively, with up to a 16.0$\times$ memory saving and a 19.0$\times$ computational cost reduction compared to the full-precision CNN. Comparing to the state-of-the-art binary models (\eg, XNOR-Net), Bi-Real net achieves 10\% higher top-1 accuracy on the 18-layer network. By using the shortcut propagating the real-valued feature map in the bottleneck structure, Bi-Real net achieves 64.5\% top-1 accuracy with an ultra-deep 152-layer structure. We also apply Bi-Real net to a real-world application, the depth estimation task. The experimental results demonstrate that a 50-layer Bi-Real net achieves superior performance over BinaryNet \cite{binarynet} and comparable accuracy to the real-valued counterpart. 

This paper extends the preliminary conference paper \cite{bi-real-net} in several aspects. 1) We generalize the idea of using the shortcut propagating the real-valued features to the ultra-deep ResNet structure with the bottleneck, enabling the application of our Bi-Real net to both shallow and deep ResNets. The idea of propagating real-valued activations in 1-bit CNNs with the shortcut is proven to be effective with the general guideline that the real-valued outputs of each convolutional layer should be added to the shortcut for propagation. 2) We propose a two-step training method targeting at the deep Bi-Real net with the bottleneck structure. By using the real-valued weights in the 3$\times$3 convolutional layers as an intermediate step, the ultra-deep Bi-Real net with the bottleneck can converge to achieve higher accuracy. 3) We conduct an ablation study on a higher-order approximation to the derivative of the sign function. We show that using the higher-order approximation rather than the piecewise linear function yields a marginal improvement in accuracy but induces computation overhead. 4) We apply our 50-layer Bi-Real net as a feature extractor on a depth estimation network and demonstrate comparable accuracy to a real-valued network.

\section{Related Work}
An overview of neural network compression and acceleration from both the hardware and software perspectives is provided in \cite{efficient-dnn-survey}. In computer vision, the network compression methods can be mainly divided into two major families. 

\noindent\textbf{Reducing the number of parameters.} Several methods have been proposed to compress neural networks by reducing the number of parameters and neural connections. Previous works on compact neural network structure design have achieved a high compression rate with a negligible accuracy degradation. In SqueezeNet \cite{squeezenet}, some 3$\times$3 convolutions are replaced with 1$\times$1 convolutions, resulting in a 50$\times$ reduction in the number of parameters. As an extreme version of Inception-v4 \cite{inception}, Xception \cite{xception} applies depth-wise separable convolution to reduce the number of parameters, which brings a memory saving as well as a speedup in convolution with a negligible accuracy drop. Based on depthwise separable convolutions, MobileNets \cite{mobilenet} builds light-weight deep neural networks and achieves a good trade-off between resource and accuracy. ResNext \cite{resnext} is proposed to use group convolution to achieve higher accuracy with a limited parameter budget. Recently, ShuffleNet \cite{shufflenet} utilizes both pointwise group convolution and channel shuffle to achieve about a 13$\times$ speedup over AlexNet with comparable accuracy.

Pruning is another effective solution for model compression and acceleration. Pruning filters \cite{filter-pruning} in the network actually removes filters together with the connected feature maps, which significantly reduces the computational costs. He \textit{et al.} \cite{channel-pruning}  pruned the channel in the network and achieved a 5$\times$ speedup with an only 0.3\% increase of errors on VGG-16. Guo \textit{et al.} \cite{dynamic-network-surgery} made on-the-fly connection pruning and efficiently compressed LeNet-5 and AlexNet by a factor of 108$\times$ and 17.7$\times$, respectively, without any accuracy loss. Li \textit{et al.} proposed to use ADMM-based method for filter pruning. In Sparse CNN \cite{sparsecnn}, a sparse matrix multiplication operation is employed to zero out more than 90\% of parameters to accelerate the learning process. Motivated by the Sparse CNN, Han \textit{et al.} proposed Deep Compression \cite{deepcompression}, which employs connection pruning, quantization with retraining, and Huffman coding to reduce the number of neural connections.

\noindent\textbf{Parameter quantization.} The study in \cite{fwfa} demonstrates that real-valued deep neural networks such as AlexNet \cite{alexnet}, GoogLeNet \cite{googlenet}, and VGG-16 \cite{vggnet} only encounter a marginal accuracy degradation when quantizing 32-bit parameters to 8-bit. In Incremental Network Quantization \cite{incremental}, Zhou \textit{et al.} quantized the parameter incrementally and showed that it is even possible to further reduce the weight precision to 2-5 bits with slightly higher accuracy on the ImageNet dataset than a full-precision network. Based on that, Zhou \textit{et al.} further proposed explicit loss-error-aware quantization \cite{elq}, which obtains comparable accuracy to the real-valued network with very low-bit parameter values. In BinaryConnect~\cite{binaryconnect}, Courbariaux \textit{et al.} more radically employed 1-bit precision weights (1 and -1) while maintaining sufficiently high accuracy on the MNIST, CIFAR10, and SVHN datasets. Ho \textit{et al.} utilized a proximal Newton algorithm with a diagonal Hessian approximation that directly minimizes the loss with regard to the binarized weights~\cite{loss-aware-binarization}. 

Quantizing weights properly can achieve considerable memory savings with little accuracy degradations. However, acceleration via weight quantization alone would be limited due to the real-valued activations (\ie, the input to convolutional layers). Several recent studies have been conducted to explore new network structures and/or training techniques for quantizing both weights and activations while minimizing the accuracy degradation. Successful attempts include DoReFa-Net \cite{dorefanet} and QNN \cite{qnn}, which explore neural networks trained with 1-bit weights and 2-bit activations, and the accuracy drops by 6.1\% and 4.9\%, respectively, on the ImageNet dataset compared to the real-valued AlexNet. Recently, Zhuang \textit{et al.} \cite{zhuangbohan} proposed to jointly train a full-precision model alongside the low-precision one, which leads to no performance decrease in a 4-bit precision network compared with its full-precision counterpart. Zhang \textit{et al.} \cite{lq-net} proposed an easy-to-train scheme of jointly training a quantized, bit-operation-compatible DNN and its associated quantizers, which can be applied to quantize weights and activations with arbitrary-bit precision. Additionally, BinaryNet \cite{binarynet} uses only 1-bit weights and 1-bit activations in a neural network and achieves comparable accuracy as a full-precision neural network on the MNIST and CIFAR-10 datasets. In XNOR-Net \cite{xnornet}, Rastegari \textit{et al.} further improved BinaryNet by multiplying the absolute mean of the weight filter and activation with the 1-bit weight and 1-bit activation to improve the accuracy. ABC-Net \cite{dji} is proposed to enhance the accuracy by using more weight bases and activation bases. The results of these studies are encouraging, but the additional usage of real-valued weights and real-valued operations offsets the memory saving and speedup of binarizing the network. In~\cite{label-refinery}, Bagherinezhad \textit{et al.} proposed label refinery technique for further improving the accuracy of the quantized networks, which is orthogonal to other quantization methods.

In this study, we aim to design 1-bit CNNs aided with a real-valued shortcut to compensate for the accuracy loss of binarization. In contrast to approaches mentioned above, adding real-valued shortcuts does not incur non-trivial real-valued operations nor extra memory. We further design 1) optimization strategies for overcoming the gradient dis-match issue and discrete optimization difficulties in 1-bit CNNs, 2) a customized initialization method, and 3) a two-step training method for the ultra-deep network. The proposed solution enables us to fully explore the potential of 1-bit CNNs with the limited resources.

\section{Methodology}
\subsection{Standard 1-Bit CNN and Its Representational Capability}
\begin{figure}[t] 
\centering
\includegraphics[width=1\linewidth]{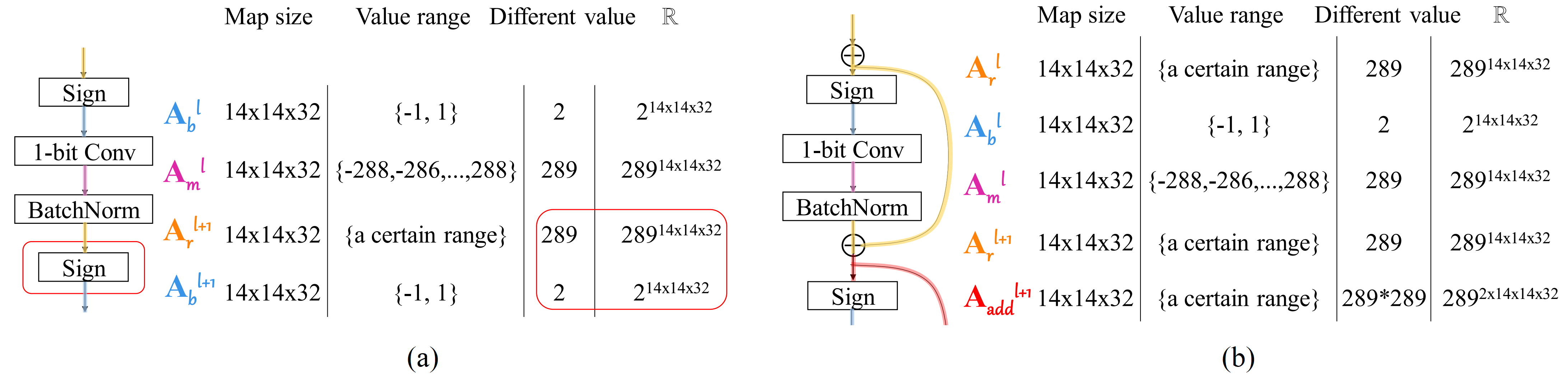}
%\vspace{-0.8cm}
\caption{The representational capability ($\mathbb{R}$) of each layer in (a) 1-bit CNNs without shortcuts, and (b) 1-bit CNNs with shortcuts. $\A_b^l$ indicates the output of the Sign function,  $\A_m^l$ denotes the output of the 1-bit convolutional layer, and $\A_r^{l+1}$ represents the output of the BatchNorm layer. The superscript $l$ indicates the block index.}
%\vspace{-0.4cm}
\label{fig:reps}
\end{figure}

1-bit convolutional neural networks (CNNs) refer to the CNN models with binary weight parameters and binary activations in intermediate convolutional layers\comment{, while the weight parameters of the first convolutional layer and the fully connected layer are still real}. 
Specifically, the binary activation and weight are obtained through a sign function, 
\begin{align}
a_b = {\rm Sign}(a_r) = \left\{  
             \begin{array}{lr}  
             - 1 & {\rm if} \ \ a_r <0 \\  
             + 1 & {\rm otherwise}  
             \end{array}  
\right. 
,
\quad \quad
w_b = {\rm Sign}(w_r) = \left\{  
             \begin{array}{lr}  
             - 1 & {\rm if} \ w_r <0 \\  
             + 1 & {\rm otherwise}  
             \end{array} 
\right. 
,
\label{eq: sign_a and sign_w}
\end{align}
where $a_r$ and $w_r$ indicate the real activation and the real weight, respectively. 
$a_r$ exists in both the training and inference processes of the 1-bit CNN, due to the convolution and batch normalization (if used). For example, given a binary activation map and a binary $3\times3$ weight kernel, the output activation could be an odd integer from $-9$ to $9$. If a batch normalization is applied, as shown in Fig. \ref{fig:reps}, then the integer activation will be transformed into real values. 
The real-valued weights will be used to update the binary weights in the training process, which will be introduced later.

Compared to real-valued CNN models with 32-bit weight parameters, the 1-bit CNNs gain up to a $32\times$ memory saving. 
Moreover, as the activation is also binary, the convolution operation could be implemented by the bitwise XNOR operation followed by a popcount operation \cite{xnornet}, \ie, 
\begin{align}
\mathbf{a_b} \cdot \mathbf{w_b}  = \text{popcount}\text{(XNOR}(\mathbf{a_b},\mathbf{w_b})), 
\label{eq: bitcount}
\end{align}
where $\mathbf{a_b}$ and $\mathbf{w_b}$ indicate the vectors of binary activations $a_{b,i}$ and binary weights $w_{b,i}$, respectively, with $i$ being the entry index. 
In contrast, the convolution operation in real-valued CNNs is implemented by the expensive real value multiplication.
Consequently, the 1-bit CNNs could obtain up to a 64$\times$ computation saving.

However, it has been demonstrated in \cite{binarynet} that the classification performance of the 1-bit CNNs is much worse than that of real-valued CNN models on large-scale datasets like ImageNet. 
We believe that the poor performance of 1-bit CNNs is caused by their low representational capacity. 
We denote $\mathbb{R}(\mathbf{x})$ as the representational capability of $\mathbf{x}$, \ie, the number of all possible configurations of $\mathbf{x}$, where $\mathbf{x}$ could be a scalar, vector, matrix, or tensor.
For example, the representational capability of 32 channels of a binary $14 \times 14$ feature map $\mathbf{A}$ is $\mathbb{R}(\mathbf{A}) = 2^{14 \times 14 \times 32} = 2^{6272}$. 
Given a $3 \times 3 \times 32$ binary weight kernel $\mathbf{W}$, each entry of $\mathbf{A} \otimes \mathbf{W}$ (\ie, the bitwise convolution output) can choose even values from (-288 to 288), as shown in Fig \ref{fig:reps}. Thus, $\mathbb{R}(\mathbf{A} \otimes \mathbf{W})$ = $289^{6272}$. Note that since the BatchNorm layer is a unique mapping, it will not increase the number of different choices but will scale the (-288,288) to a particular value. If adding the sign function behind the output, each entry in the feature map is binarized, and the representational capability shrinks to $2^{6272}$ again.

\subsection{Shallow Bi-Real Net Model and Its Representational Capability}
\label{sec:1-layer-per-block-explanation}
We propose to preserve the real activations before the sign function to increase the representational capability of the 1-bit CNN through a simple shortcut.  
Specifically, as shown in Fig. \ref{fig:reps}(b), one block indicates the structure 
"Sign $\rightarrow$ 1-bit convolution $\rightarrow$ batch normalization $\rightarrow$ addition operator". 
The shortcut connects the input activations to the sign function in the current block to the output activations after the batch normalization in the same block, and these two activations are added through an addition operator, and then the combined activations are input to the sign function in the next block. 
The representational capability of each entry in the added activations is $289^2$. 
Consequently, the representational capability of each block in the 1-bit CNN with the above shortcut becomes $(289^2)^{6272}$. 
As both real and binary activations are retained, we call the proposed model Bi-Real net.

The representational capability of each block in the 1-bit CNN is significantly enhanced due to the simple identity shortcut. 
The only additional cost of computation is the addition operation of two real activations, as these real activations already exist in the standard 1-bit CNN (\ie, without shortcuts). Moreover, as the activations are computed on the fly, no additional memory is needed. 

\begin{figure}
\centering
\includegraphics[width=0.6\linewidth]{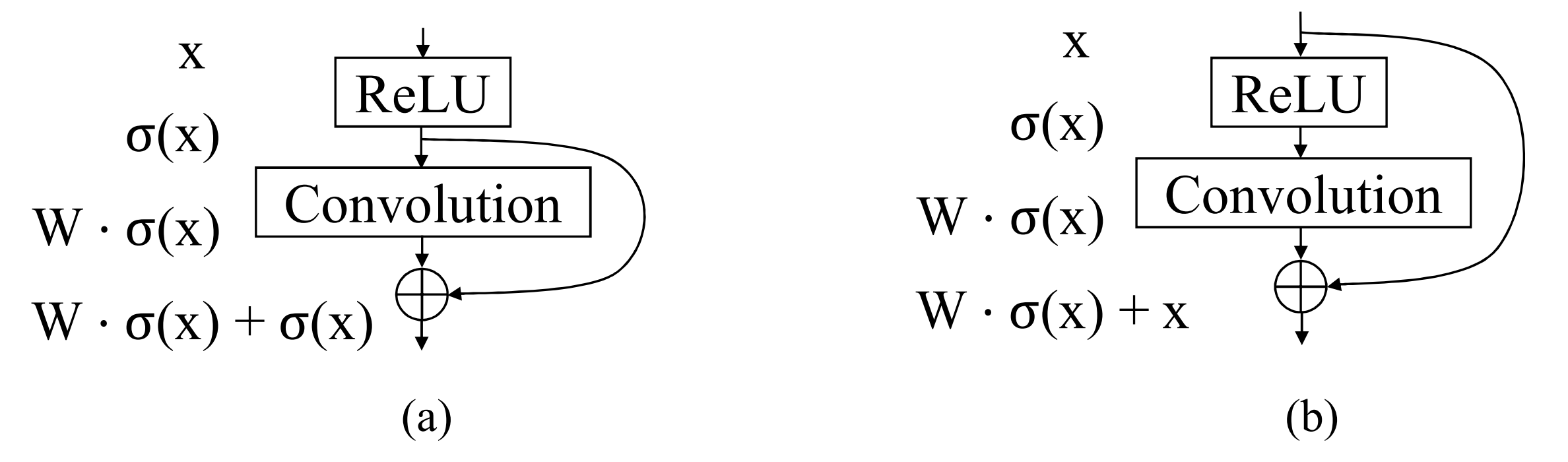}
\caption{ (a) The 1-layer-per-block structure that will not work \cite{resnet}. (b) The proposed Bi-Real net basic block with non-linearity inside that does work.}
\label{fig:1-layer-per-block-analysis}
\end{figure}

It is mentioned in ResNet \cite{resnet} that a residual block with only one convolutional layer will lose the superiority of the shortcut connections. As shown in Fig. \ref{fig:1-layer-per-block-analysis} (a), $W \cdot \sigma(x) + \sigma(x) = (W+1) \cdot \sigma(x)$, which means the identity mapping can be learned by the weight matrix and the block will act as a plain layer. However, He \textit{et al.}, in \cite{identity_mapping}, proposed to move the nonlinear function into the block. Based on this, we find that 1-layer-per-block structure with a nonlinear function inside the block performs residual learning and is distinct from a plain layer. As shown in Fig. \ref{fig:1-layer-per-block-analysis} (b), $W \cdot \sigma(x) + x \neq (W+1) \cdot \sigma(x)$. Thus, Bi-Real net with the shortcut connecting every layer's output extensively utilizes the shortcut and this so-called 1-layer-per-block design brings a huge benefit for 1-bit CNNs.

\begin{figure}[t]
\centering
\includegraphics[width=\linewidth]{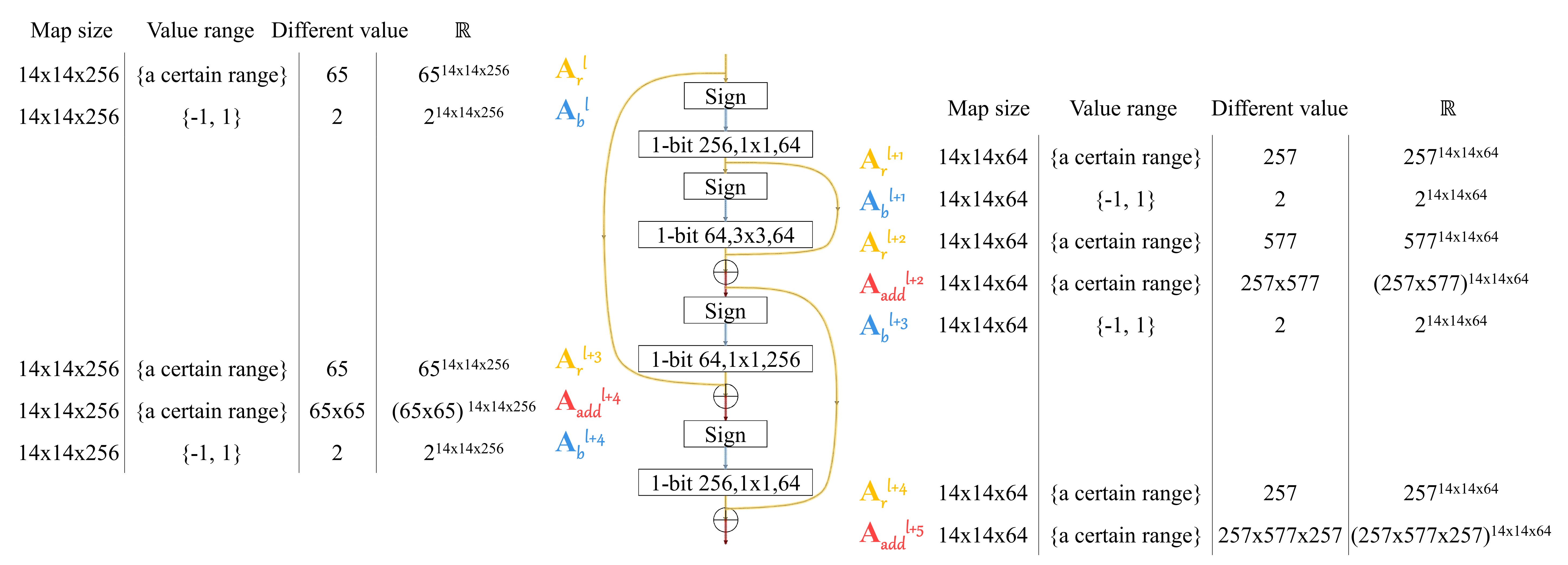}
\caption{The representational capability ($\mathbb{R}$) of each layer in the deep Bi-Real net model. $\A_b$ indicates the binary activations;  $\A_r$ represents the real-valued activations; $\A_{add}$ denotes the real-valued activations after element-wise add operation; superscript $l$ indicates the block index.}
\label{fig:representation-deep}
\end{figure}

\subsection{Deep Bi-Real Net Model with the Bottleneck and Its Representational Capability} 
Binarizing an ultra-deep network is not trivial. Increased depth induces a training difficulty and also raises a higher requirement for the network structure design. As ResNet is the most prevalent network backbone structure, we propose to binarize the deep ResNet with the bottleneck structure to verify the superiority of our shortcut design and the training algorithm on a deep network.

Although a deep ResNet bottleneck structure already has a shortcut for each block, binarizing the convolutional layers would discard the real-valued outputs of the intermediate layers in a bottleneck block. Thus, we suspect that it could be much more effective to use the shortcut to propagate all the real-valued outputs of each 1-bit convolutional layer. Based on this conjecture, we propose to use another shortcut to propagate real-valued features generated inside the bottleneck. As shown in Fig. \ref{fig:representation-deep}, the newly added shortcut connects the input activations to the sign functions before the 3$\times$3 convolutional layers and the output activations of the 3$\times$3 convolutional layers in series, by an adding operation. This shortcut path together with the original shortcut path jointly collects all the real-valued outputs of the convolutional layers, greatly enhancing the representational capability. As illustrated in Fig. \ref{fig:representation-deep}, the representational capability of each entry in the original shortcut path, \ie, left path, grows from 65 to 65 $\times$ 65 with the added shortcut. The representational capability of each entry in the additional shortcut path, \ie, right path, grows from 257 to 257 $\times$ 577 $\times$ 257. The representational capability of each entry after the adding operation will be the product of those of the input entries, which grows exponentially with the network depth and greatly contributes to the final accuracy.

\begin{figure}[t]
\centering
\includegraphics[width=0.75\linewidth]{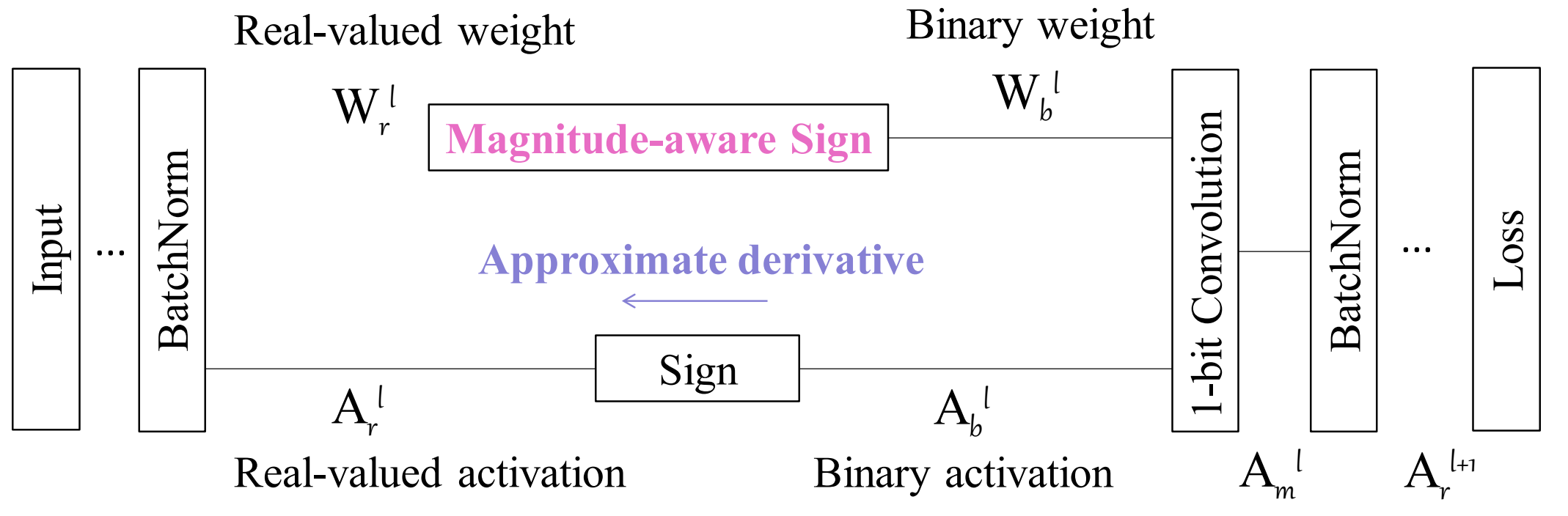}
\caption{An illustration of the training process of the 1-bit CNNs, with $A$ being the activation, $W$ being the weight, and superscript $l$ denoting the $l^{\textit{th}}$ block consisting with Sign, 1-bit Convolution, and BatchNorm. Subscript $r$ denotes real value, $b$ denotes binary value, and $m$ denotes the intermediate output before the BatchNorm layer.}
\label{fig:training}
\end{figure}

\subsection{Training Bi-Real Net}
As both activations and weight parameters are binary, the continuous optimization method, \ie, the stochastic gradient descent (SGD), cannot be directly adopted to train the 1-bit CNN. 
There are two major challenges. One is how to compute the gradient of the sign function on activations, which is non-differentiable, while the other is that the gradient of the loss with respect to the binary weight is too small to change the sign of the weight. 
The authors of \cite{binarynet} proposed to adjust the standard SGD algorithm to approximately train the 1-bit CNN. Specifically, the gradient of the sign function on activations is approximated by the gradient of the piecewise linear function, as shown in Fig. \ref{fig:activation_back}(b). To tackle the second challenge, the method proposed in \cite{binarynet} updates the real-valued weights by the gradient computed with regard to the binary weight, and obtains the binary weight by taking the sign of the real weights. 
As the identity shortcut will not add difficulty for training, the training algorithm proposed in \cite{binarynet} can also be adopted to train the Bi-Real net model. 
However, we propose a novel training algorithm to tackle the above two major challenges, which is more suitable for the Bi-Real net model as well as other 1-bit CNNs. 
Additionally, we also propose a novel initialization method.

We present a graphical illustration of the training of Bi-Real net in Fig. \ref{fig:training}. The identity shortcut is omitted in the graph for clarity, as it will not change the main part of the training algorithm. 

\subsubsection{Approximation to the derivative of the sign function with respect to activations.}
\begin{figure}[t]
\centering
\includegraphics[width=12cm]{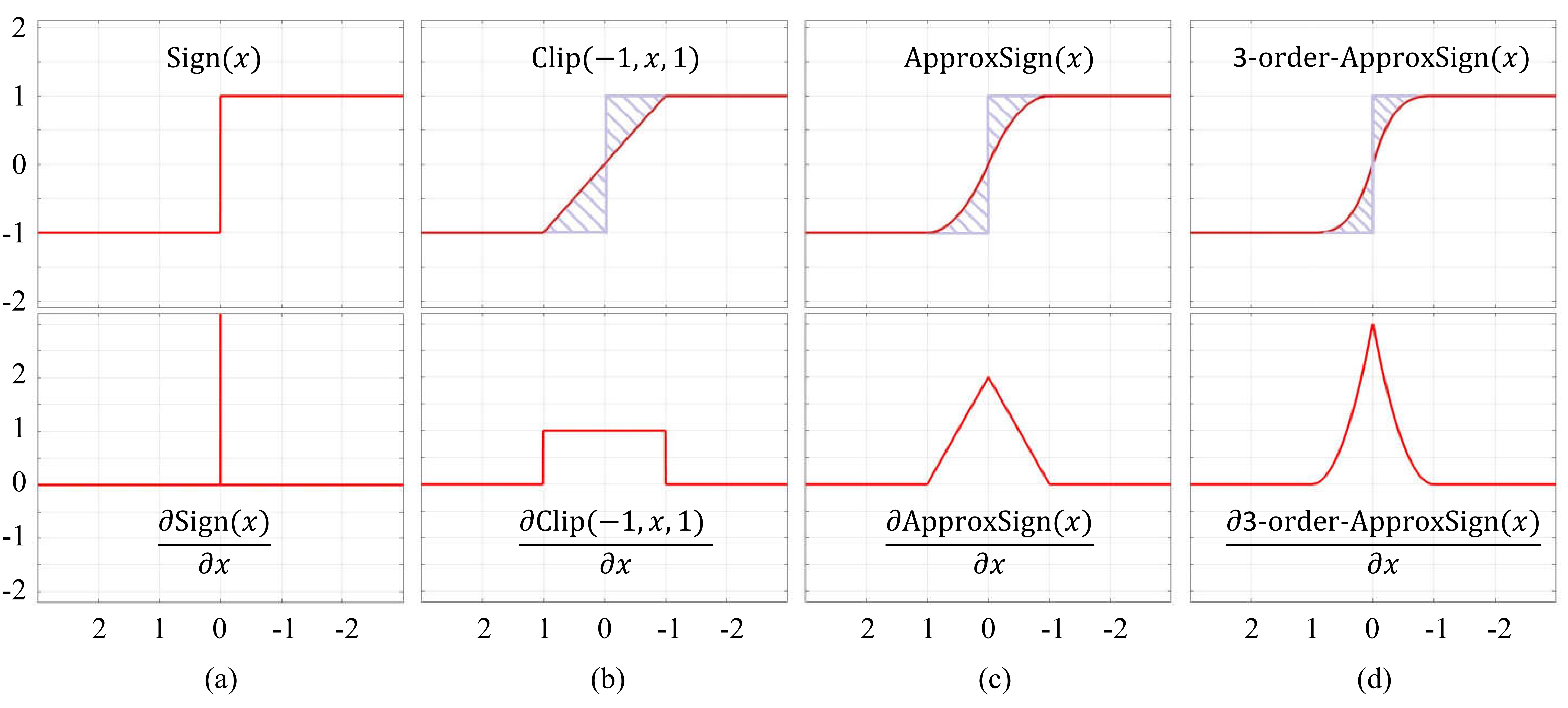}
\caption{(a) The sign function and its derivative. (b) The clip function and its derivative for approximating the derivative of the sign function, as proposed in \cite{binarynet}. (c) The proposed differentiable piecewise polynomial function and its triangle-shaped derivative for approximating the derivative of the sign function in gradients computation, and (d) the differentiable third-order piecewise polynomial function and its derivative.} 
\label{fig:activation_back}
\end{figure}

As shown in Fig. \ref{fig:activation_back}(a), the derivative of the sign function is an impulse function, which cannot be utilized in training. 
\label{sec:activation_back}

\begin{flalign}
\frac{\partial \mathcal{L}}{\partial \mathbf{A}_r^{l,t}} 
= 
\frac{\partial \mathcal{L}}{\partial \mathbf{A}_b^{l,t}}  
\frac{\partial \mathbf{A}_b^{l,t}}{\partial \mathbf{A}_r^{l,t}}
=
\frac{\partial \mathcal{L}}{\partial \mathbf{A}_b^{l,t}}  
\frac{\partial Sign(\mathbf{A}_r^{l,t})}{\partial \mathbf{A}_r^{l,t}}
\approx
\frac{\partial \mathcal{L}}{\partial \mathbf{A}_b^{l,t}}  
\frac{\partial F(\mathbf{A}_r^{l,t})}{\partial \mathbf{A}_r^{l,t}},
\label{eq: derivative wrt A_r}
\end{flalign}
where $F(\mathbf{A}_r^{l,t})$ is a differentiable approximation of the non-differentiable $Sign(\mathbf{A}_r^{l,t})$. 
In \cite{binarynet},  $F(\mathbf{A}_r^{l,t})$ is set as the clip function, leading to the derivative as a step-function (see Fig. \ref{fig:activation_back}(b)).
In this work, we utilize a piecewise polynomial function (see Fig. \ref{fig:activation_back}(c)) as the approximation function, as follows:

\begin{align}
\label{eq4}
F(a_r) = \left\{  
             \begin{array}{lr}  
             - 1 & {\rm if} \ a_r  < -1 \\ 
             2a_r+a_r^2 \ \ &{\rm if} -1 \leqslant a_r < 0 \\
             2a_r-a_r^2 &{\rm if} \ 0 \leqslant a_r < 1 \\
             1 & {\rm otherwise}  
             \end{array} 
\right. 
.
\end{align}

\noindent
As shown in Fig. \ref{fig:activation_back}, the shaded areas with blue slashes reflect the difference between the sign function and its approximation. The shaded area corresponding to the clip function is $1$, while that corresponding to Eq. \eqref{eq4} is $\frac{2}{3}$. 
We conclude that Eq. \eqref{eq4} is a closer approximation to the sign function than the clip function.
Consequently, the derivative of Eq. \eqref{eq4} is formulated as 
\begin{align}
\label{eq4-2}
\frac{\partial F(a_r)}{\partial a_r} = \left\{  
             \begin{array}{lr}  
             2+2a_r \ \ \ \ &{\rm if} \ -1 \leqslant a_r < 0 \\
             2-2a_r &{\rm if} \ 0 \leqslant a_r < 1 \\
             0 & {\rm otherwise}  
             \end{array} 
\right. 
,
\end{align}
which is a piecewise linear function. 

Although we can use a higher-order approximation to obtain closer approximation, the gain would be limited. We carry out an ablation study on the third-order approximation to the sign function, formulated as 

\begin{align}
\label{eq5}
F(a_r) = \left\{  
             \begin{array}{lr}  
             - 1 & {\rm if} \ a_r  < -1 \\ 
             (a_r+1)^3-1 \ \ &{\rm if} -1 \leqslant a_r < 0 \\
             (a_r-1)^3+1 &{\rm if} \ 0 \leqslant a_r < 1 \\
             1 & {\rm otherwise}  
             \end{array} 
\right. 
.
\end{align}
Its derivative is a piecewise quadratic function:
\begin{align}
\label{eq6}
\frac{\partial F(a_r)}{\partial a_r} = \left\{  
             \begin{array}{lr}  
             3(a_r+1)^2 \ \ &{\rm if} -1 \leqslant a_r < 0 \\
             3(a_r-1)^2 &{\rm if} \ 0 \leqslant a_r < 1 \\
             0 & {\rm otherwise}  
             \end{array} 
\right. 
.
\end{align}

Intuitively, the gradient dis-match decreasing from the second-order approximation to the third-order approximation is small, as shown in Fig. \ref{fig:activation_back}, and the experimental results also show that the corresponding accuracy increase is very limited. Thus, we conclude that the second-order approximation is sufficient. The limited gain from a higher-order approximation is not worthwhile for the computational overhead.

\subsubsection{Magnitude-aware binarization with respect to weights.}
Updating the binary parameters is challenging. The naive solution induces the magnitude dis-match issue between the binary weights and real-valued weights. This magnitude dis-match issue is in turn aggravated in gradient with the existence of the BatchNorm layer and hampers the convergence of 1-bit networks. To tackle this we use a magnitude-aware binarization scheme to match the magnitudes between the binary weights and the real-valued weights at the training time. After training, we use the naive sign function for weight binarization to keep inference simple and fast.

Here we present how to update the binary weight parameter in the $l^{th}$  block.
The standard gradient descent algorithm cannot be directly applied as the gradient is not large enough to change the binary weights.

To tackle this problem, the method of \cite{binarynet} introduces a real weight $\mathbf{W}_r$ and a sign function during training. Hence the binary weight parameter can be seen as the output to the sign function, \ie,  $\mathbf{W}_b = Sign(\mathbf{W}_r)$, as shown in the upper sub-figure in Fig. \ref{fig:training}. 
$\mathbf{W}_r$ is updated using gradient calculated with respect to $\mathbf{W}_b$ in the backward pass, as follows:
\begin{align}
\mathbf{W}_r^{t+1} = \mathbf{W}_r^{t} - \eta \frac{\partial \mathcal{L}}{\partial \mathbf{W}_b^{t}} \cdot \textbf{1}_{|\mathbf{W}_r^{t}|<1}.
\label{eq: update for W_r}
\end{align}

\begin{figure}[t]
\centering
\includegraphics[width=\linewidth]{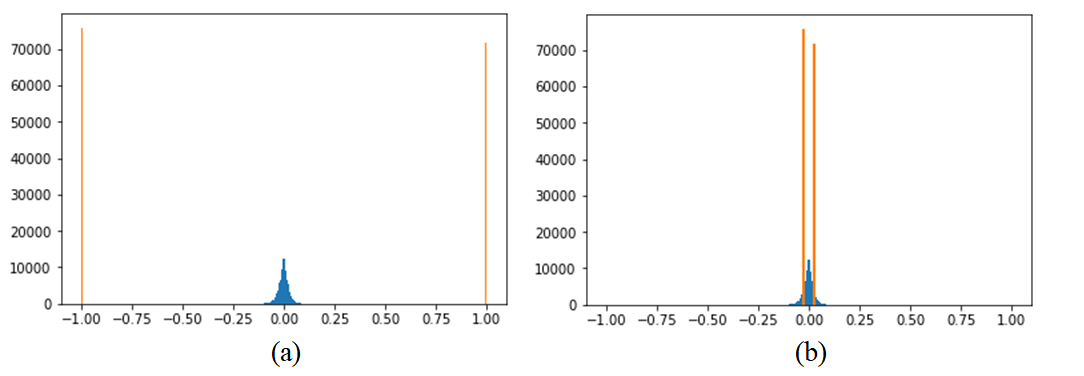}\caption{The weight distribution in the 18-layer Bi-Real net, with the blue distribution denoting real-valued weights and the yellow line denoting binarized weights, respectively. (a) Magnitude difference between real-valued weights and the sign of real-valued weights (\ie, binarized weights); (b) Magnitude difference between real-valued weights and binarized weights obtained by the magnitude-aware sign.}
\label{fig:weight_distribution}
\end{figure}

This method solves the problem of updating the discrete binary weights, but the sign function binarizing weights induces a magnitude dis-match problem between binary weights and real-valued weights. As shown in Fig. \ref{fig:weight_distribution} (a), the magnitude of binary weights $\parallel \W_b\parallel_{1,1}$ equals to 1; however, for empirical reasons, the magnitude of real weights $\parallel \W_r\parallel_{1,1}$ is around 0.1.

Thus in binary networks, we have 
\begin{align}
\label{eq:binaryweight1}
\parallel \W_b\parallel_{1,1} = \alpha \parallel \W_r\parallel_{1,1},
\end{align}

\noindent where $\alpha \approx 10$ is a non-negligible number.

With the existence of the BatchNorm layer in the prevalent architectures, the magnitude difference between $\mathbf{W}_b$ and $\mathbf{W}_r$ will incur a reciprocal magnitude difference in gradient, which in turn harms the convergence of the binary networks. We explain this phenomenon through starting with a simple lemma.

\textbf{\textit{Lemma1:}} With the presence of a BatchNorm layer in a convolutional neural network, if every element in the weight kernel is amplified by $\alpha$, the gradient will become $\frac{1}{\alpha}$ of the previous gradient.

\textbf{\textit{Proof:}} For clarity, we assume that there is only one weight kernel, \ie, $\mathbf{W}$ is a matrix. 

\textit{Forward Pass:}
We consider a convolutional layer followed by a BatchNorm layer in the convolutional neural network, where the original parameters are denoted with superscript (1) and the parameters after scaling are denoted with superscript (2).
When weights in the convolutional layer are amplified by $\alpha$,

\begin{align}
\label{eq:mw1}
\textbf{W}^{(2)} = \alpha \textbf{W}^{(1)},
\end{align}

\noindent and the output of the convolutional layer is 
\begin{align}
\label{eq:mw2}
\textbf{Y} = \textbf{X} \cdot \textbf{W}, \ \ so, \ \ \textbf{Y}^{(2)} = \alpha \textbf{Y}^{(1)}.
\end{align}

For the BatchNorm layer which normalizes $\textbf{Y}$, the mean ($\mu$) and variance ($\sigma$) can be calculated as
\begin{align}
\label{eq:mw4}
& \mu = \frac{1}{m} \sum^m_{i=1}Y_i , \ \ \ \sigma^2 = \frac{1}{m}\sum^m_{i=1}(Y_i-\mu)^2 ,
\end{align}

\noindent where \textit{m} is the number of entries in the weight matrix. Thus, we have 
\begin{align}
\label{eq:mw5}
&\mu^{(2)} = \alpha \mu^{(1)}, \ \ \ \sigma^{(2)} = \alpha \sigma^{(1)}.
\end{align}

The output $\textbf{Z}$ of the BatchNorm layer in the forward pass is independent of the scaling factor $\alpha$,

\begin{align}
\label{eq:mw6}
&\textbf{Z} = \frac{\textbf{Y}-\mu}{\sigma}, \ \ so, \ \ \textbf{Z}^{(2)} = \textbf{Z}^{(1)} .
\end{align}

\textit{Backward Pass:} As we assume that other layers besides the concerned layer are the same, the gradient back propagated to the corresponding BatchNorm layer should be identical, that is,

\begin{align}
\label{eq:mw7}
\frac{\partial\mathcal{L}}{\partial \textbf{Z}^{(2)}} = \frac{\partial\mathcal{L}}{\partial \textbf{Z}^{(1)}} .
\end{align}

We can calculate the gradient with respect to the weights by Chain Rule:

\begin{align}
\label{eq:mw8}
&\frac{\partial\mathcal{L}}{\partial \textbf{W}} = \frac{\partial\mathcal{L}}{\partial \textbf{Z}} \cdot \frac{\partial \textbf{Z}}{\partial \textbf{Y}} \cdot \frac{\partial \textbf{Y}}{\partial \textbf{W}} = \frac{\partial\mathcal{L}}{\partial \textbf{Z}} \cdot \frac{1}{\sigma} \cdot \textbf{X} .
\end{align}

Since $\sigma^{(2)} = \alpha \sigma^{(1)}$, we have

\begin{align}
\label{eq:mw10}
\frac{\partial\mathcal{L}}{\partial \textbf{W}^{(2)}} = \frac{1}{\alpha} \frac{\partial\mathcal{L}}{\partial \textbf{W}^{(1)}} .
\end{align}

The gradient with respect to the weight is scaled to $\frac{1}{\alpha}$.
Obviously this lemma holds in the convolutional layer with multiple output channels.

Proof completed.

According to \textbf{\textit{Lemma1}}, this magnitude difference between the binary weights and the real-valued weights induces the gradient to be rescaled in the converse direction:
\begin{align}
\label{eq:binaryweight2}
\parallel\frac{\partial\mathcal{L}}{\partial \textbf{W}_b}\parallel_{1,1} \approx \frac{1}{\alpha} \parallel\frac{\partial\mathcal{L}}{\partial \textbf{W}_r}\parallel_{1,1}.
\end{align}

This effect makes the weight update of the weights in the binary network in-precise and hinders the binary network from converging to a higher accuracy. 
To address this issue, we adopt the magnitude-aware binarization $\mathbf{W}_b =  || \W_r ||_{1,1} \times  sign(\textbf{W}_r)$ for matching the dimension between binary weight and real-valued weights. Thus, we have 

\begin{align}
\label{eq:binaryweight3}
\parallel \W_b\parallel_{1,1} = \parallel \W_r\parallel_{1,1} \ \  and \ \ \parallel\frac{\partial\mathcal{L}}{\partial \textbf{W}_b}\parallel_{1,1} \approx \parallel\frac{\partial\mathcal{L}}{\partial \textbf{W}_r}\parallel_{1,1} .
\end{align}

After training the binary network with the more accurate gradients, we no longer need to re-scale the binary weights at inference time, as the output of the BatchNorm layer is independent of the scaling factor. By setting $\textbf{W}_b' = sign(\W_r) $ and $\mu' = \frac{\mu}{\parallel \W_r\parallel_{1,1}}, \sigma' = \frac{\sigma}{\parallel \W_r\parallel_{1,1}}$, we can obtain the same value in the output,

\begin{align}
\label{eq:binaryweight3}
\textbf{Z}' = \frac{\textbf{Y}_b' - \mu_b'}{\sigma_b'} = \frac{\textbf{X}' \cdot \textbf{W}_b' - \mu_b'}{\sigma_b'} = \frac{\textbf{X}\cdot \frac{\textbf{W}_b}{\parallel\W_r\parallel_{1,1}}-\frac{\mu_b}{\parallel \W_r\parallel_{1,1}}}{\frac{\sigma_b}{\parallel \W_r\parallel_{1,1}}} = \textbf{Z}.
\end{align}

Thus we can simply use the sign function for binarizing the weights at the inference phase for easy deployment.

Using $\mathbf{W}_b = \parallel \W_r\parallel_{1,1} \cdot sign(\textbf{W}_r)$ for weight binarization is first proposed in XNOR-Net \cite{xnornet} and inherited by the following works~\cite{dji,train_binary}. But previous works use this scaling factor for enhancing the representational capability of the binary weights in both training time and inference time, which results in extra computation in deploying the binary model. To the best of our knowledge, we are the first to explicitly point out that this scaling factor can be used as an auxiliary parameter to help convergence at training time and be normalized by the BatchNorm layer at inference time.

\subsubsection{Initialization.}
\begin{figure}[t]
\centering
\includegraphics[width=\linewidth]{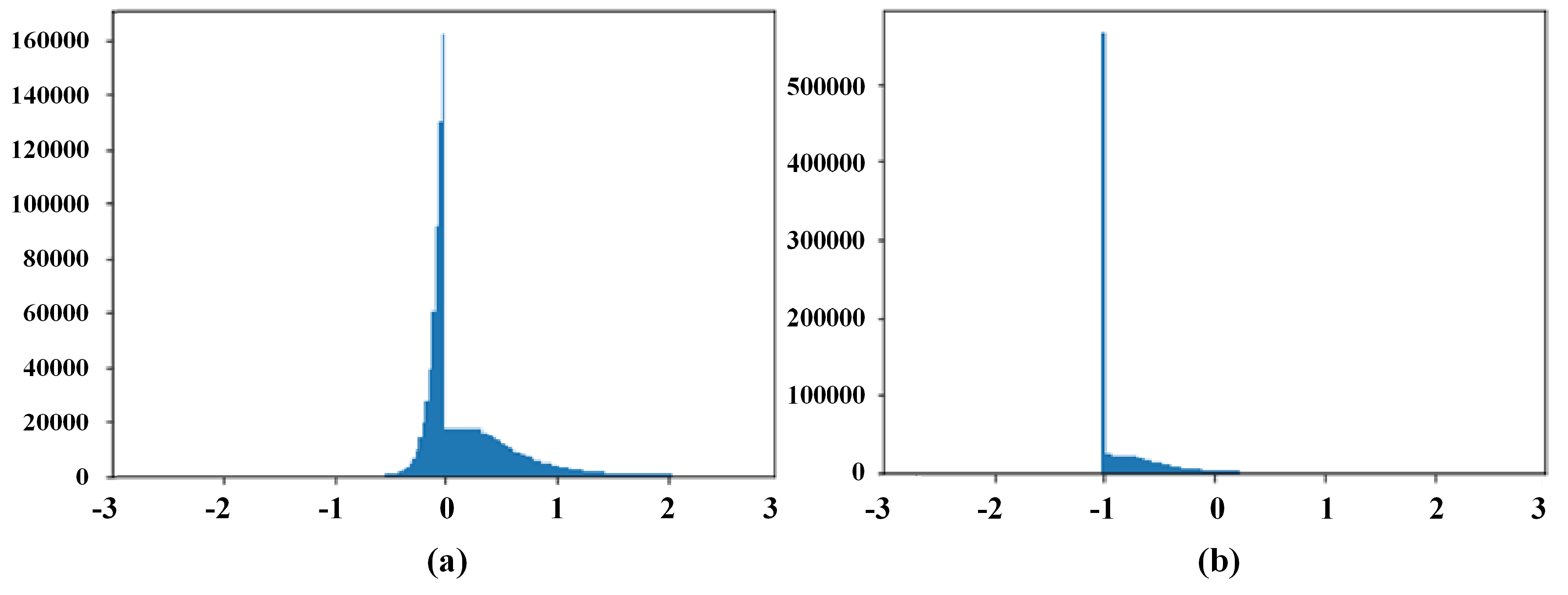}
\caption{The activation distribution in the 18-layer Bi-Real net with different non-linearity functions: (a) ReLU function, and (b) clip function. }
\label{fig:activation_distribution}
\end{figure}

As discussed in the previous section, training the 1-bit CNN means updating the stored real-valued weight and using its sign to update the binary weight. The value of a real-valued weight denoting how likely it is a binary weight is going to change its sign. A good initialization of the real-valued weights is of great importance for a rapid convergence and a high accuracy of the model. Previous works proposed to fine-tune the 1-bit CNN from the corresponding real-valued network with ReLU non-linearity \cite{dji}. However, the activation of ReLU is non-negative, as shown in Fig. \ref{fig:activation_distribution} (a), while that of Sign is $-1$ or $+1$. Due to this difference, the real-valued CNNs with ReLU may not provide a suitable initial point for training the 1-bit CNNs. Instead, we propose to replace ReLU with $\text{clip}(-1,x,1)$ to pre-train the real-valued CNN model. The activation of the clip function is closer to the sign function than ReLU, as shown in Fig. \ref{fig:activation_distribution} (b), and yields a better initialization which will be further validated in the following experiments.

The initialization method and the magnitude-aware binarization with respect to weights can be viewed as a whole process to match the magnitude of activations and weights in the real-valued network with that in the 1-bit CNNs. Replacing the ReLU non-linearity with the clip function in initialization reshapes the activation distribution to be closer to -1 and +1. Then the magnitude-aware sign matches the magnitude of the binary weights and the real-valued weights to obtain more effective gradients. After training for convergence, we use the BatchNorm layer to normalize the scaling factor and scale back the binary weights to -1 and +1. In this process, we manage to convert the real-valued networks to 1-bit CNNs.

\subsubsection{Two-step training method for a deep 1-bit CNN with the bottleneck structure.} 
To further alleviate the difficulty in training a deep 1-bit CNN, we followed the progressive quantization methods~\cite{zhuangbohan} to binarize the entire network in a two-step manner, in which the real-valued weights in the 3$\times$3 convolutional layers are used as auxiliary variables. Specifically, we first binarize the activations and weights in all the 1$\times$1 convolutional layers and the activations in the 3$\times$3 convolutional layers in the bottleneck blocks. We train the network till convergence, and then binarize the weights in the 3$\times$3 convolutional layers. With the proposed two-step training method, we manage to decompose the big challenge of binarizing the deep network into two sub-problems, facilitating the network converge to higher accuracy.

\section{Experiments}       
In this section, we first introduce the dataset for experiments, and present implementation details in Sec \ref{sec:dataset_implementation}. Then, we conduct an ablation study in Sec. \ref{sec:ablation_study} to investigate the effectiveness of the proposed techniques. This is followed by comparing our Bi-Real net with other state-of-the-art binary networks regarding accuracy in Sec \ref{sec:accuracy_comparison}. Sec. \ref{sec:efficiency_comparison} reports the memory usage and computation cost in comparison with other networks. In Sec. \ref{sec:depth_estimation}, we deploy the Bi-Real net to a real-world application, the depth estimation task. 

\subsection{Dataset and Implementation Details}
\label{sec:dataset_implementation}
The experiments are carried out on the ILSVRC12 ImageNet classification dataset \cite{imagenet}. ImageNet is a large-scale dataset with 1.2 million training images and 50K validation images of 1,000 classes. Compared to other datasets like CIFAR-10 \cite{cifar10} or MNIST \cite{mnist}, ImageNet is more challenging due to its large scale and great diversity. The study on this dataset will validate the superiority of the proposed Bi-Real network structure and the effectiveness of the novel training techniques for 1-bit CNNs. In our experiment, we report both the top-1 and top-5 accuracies.

For each image in the ImageNet dataset, the lower dimension of the image is rescaled to 256 while keeping the aspect ratio intact. For \textit{training}, a random crop of size 224 $\times$ 224 is selected from the rescaled image or its horizontal flip. \comment{Note that, in contrast to XNOR-Net and the full-precision ResNet, we do not use the operation of random resize, which might improve the performance further.} For \textit{inference}, we employ the 224 $\times$ 224 center crop from images. 

\noindent\textbf{Pre-training:} We fine-tune the real-valued network with the clip non-linear function from the corresponding network with the ReLU non-linear function. For easy convergence, we use the network with the leaky clip as a transition, which has a negative slope of 0.1 instead of 0 outside the range of (-1,1). The fine-tuning procedure is ReLU $\rightarrow$ Leaky-Clip $\rightarrow$ Clip.

\noindent\textbf{Training:} We train four instances of the Bi-Real net, including an \textit{18-layer}, a \textit{34-layer}, a \textit{50-layer}, and a \textit{152-layer Bi-Real net}.  The training of them consists of two steps: training the 1-bit convolutional layer and retraining the BatchNorm. In the first step, the weights in the 1-bit convolutional layer are binarized using the magnitude-aware binarization with respect to the weights. We use the SGD solver with the momentum of 0.9 and set the weight-decay to 0, which means that we no longer encourage the weights to be close to 0. For the 18-layer Bi-Real net, we run the training algorithm for 20 epochs with a batch size of 128. The learning rate starts at 0.01 and is decayed twice by multiplying 0.1 at the 10\textit{th} and the 15\textit{th} epoch. For the 34-layer Bi-Real net, the training process includes 40 epochs and the batch size is set to 1024. The learning rate starts at 0.08 and is multiplied by 0.1 at the 20\textit{th} and the 30\textit{th} epoch respectively. For the 50-layer Bi-Real net and 152-layer Bi-Real net, the first step is further divided into two sub-steps: 1) binarizing the activations and weights in 1$\times$1 convolutional layers along with the activations in 3$\times$3 convolutional layers, and 2) binarizing the weights in 3$\times$3 convolutional layers. For the 50-layer Bi-Real net, each sub-step includes 100 epochs and the batch size is set to 800. The learning rate starts from 0.064 and is multiplied by 0.1 at the 50\textit{th} and the 75\textit{th} epoch respectively. For the 152-layer Bi-Real net, each sub-step includes 120 epochs and the batch size is set to 4000. The learning rate starts at 0.1 and is multiplied by 0.1 at the 60\textit{th} and the 90\textit{th} epoch, respectively. In the second step, we constrain the weights to -1 and 1, and set the learning rate in all convolutional layers to 0 and retrain the BatchNorm layer for one epoch to absorb the scaling factor.

\noindent\textbf{Inference:} we use the trained model with binary weights and binary activations in the 1-bit convolutional layers for inference.

\begin{figure}[t]
\label{fig:three_building_blocks}
\centering
\includegraphics[height=2.5cm]{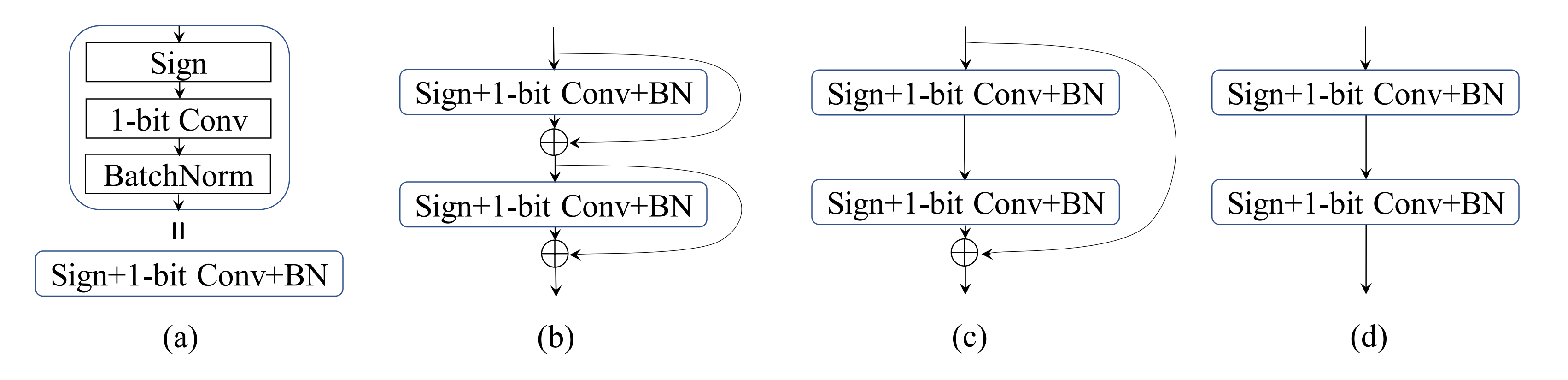}
\caption{(a) The elemental block structure of  conjoint layers of Sign, 1-bit Convolution, and the BatchNorm. Three networks differ in the shortcut design of connecting the blocks, shown in (b), (c) and (d). (b) Bi-Real net with shortcut bypassing every block, (c) Res-Net with shortcut bypassing two blocks, which corresponds to the ReLU-only pre-activation proposed in \cite{identity_mapping}, and (d) Plain-Net without the shortcut. These three structures shown in (b), (c) and (d) have the same number of weights.}
\label{fig:compare_structure}
\end{figure}

\subsection{Ablation Study}
\label{sec:ablation_study}
\noindent\textbf{Three building blocks.} The shortcut in our Bi-Real net transfers real-valued representation without additional memory cost, which plays an important role in improving its capability. To verify its importance, we implemented a Plain-Net structure without shortcuts, as shown in Fig. \ref{fig:compare_structure} (d), for comparison. At the same time, as our network structure employs the same number of weight filters and layers as the standard ResNet, we also carry out a comparison with the standard ResNet shown in Fig. \ref{fig:compare_structure} (c). For fair comparison, we adopt the ReLU-only pre-activation ResNet structure in \cite{identity_mapping}, which differs from Bi-Real net only in the structure of two layers per block instead of one layer per block. The layer order and shortcut design in Fig. \ref{fig:compare_structure} (c) are also applicable for 1-bit CNNs. The comparison can justify the benefit of implementing our Bi-Real net by specifically replacing the 2-conv-layer-per-block Res-Net structure with two 1-conv-layer-per-block Bi-Real structures.

As discussed in Sec. 3, we propose to overcome the optimization challenges induced by discrete weights and activations by 1) approximation to the derivative of the sign function with respect to activations, 2) magnitude-aware binarization with respect to weights, and 3) clip initialization. To study how these proposals benefit the 1-bit CNNs individually and collectively, we train the 18-layer structure and the 34-layer structure with a combination of these techniques on the ImageNet dataset. Thus, we derive $2 \times 3 \times 2 \time 2 \times 2 \times 2= 48$ pairs of values of top-1 and top-5 accuracies, which are presented in Table \ref{table:ablation study}.

\setlength{\tabcolsep}{1pt}
\begin{table}[t]
\scriptsize
\begin{center}
\caption{Top-1 and top-5 accuracies (in percentage) of different combinations of the three proposed techniques on the three different network structures, Bi-Real net, ResNet and Plain Net, shown in Fig. \ref{fig:compare_structure}.}
\label{table:ablation study}
%\vspace{-0.2cm}
\begin{tabular}{lllllllllllllll}
\hline
\noalign{\smallskip}
Initiali- & Weight &Activation & \multicolumn{2}{c}{Bi-Real-18} &  \multicolumn{2}{c}{Res-18} & \multicolumn{2}{c}{Plain-18} &\multicolumn{2}{c}{Bi-Real-34} &\multicolumn{2}{c}{Res-34} &\multicolumn{2}{c}{Plain-34}  \\
zation&update&backward&top-1&top-5&top-1&top-5&top-1&top-5&top-1&top-5&top-1&top-5&top-1&top-5\\
\noalign{\smallskip}
\hline
\noalign{\smallskip}
\multirow{5}{*}{ReLU} & \multirow{2}{*}{Original} &Original&\cellcolor{Gray}32.9& \cellcolor{Gray}56.7& 27.8& 50.5& 3.3& 9.5& \cellcolor{Gray}53.1& \cellcolor{Gray}76.9& 27.5& 49.9& 1.4& 4.8\\
\cline{3-15}
\noalign{\smallskip}
& &Proposed&\cellcolor{Gray}36.8& \cellcolor{Gray}60.8& 32.2& 56.0& 4.7& 13.7& \cellcolor{Gray}58.0& \cellcolor{Gray}81.0& 33.9& 57.9& 1.6& 5.3\\
\cline{2-15}
\noalign{\smallskip}
&\multirow{2}{*}{Proposed} &Original&\cellcolor{Gray}40.5& \cellcolor{Gray}65.1& 33.9& 58.1& 4.3& 12.2& \cellcolor{Gray}59.9& \cellcolor{Gray}82.0& 33.6& 57.9& 1.8& 6.1\\
\cline{3-15}
\noalign{\smallskip}
&&Proposed&\cellcolor{Gray}47.5& \cellcolor{Gray}71.9& 41.6& 66.4& 8.5& 21.5& \cellcolor{Gray}61.4& \cellcolor{Gray}83.3& 47.5& 72.0& 2.1& 6.8\\
\cline{2-15}
\noalign{\smallskip}
&\multicolumn{2}{l}{Real-valued Net} &\cellcolor{Gray}68.5& \cellcolor{Gray}88.3& 67.8& 87.8& 67.5& 87.5& \cellcolor{Gray}70.4& \cellcolor{Gray}89.3& 69.1& 88.3& 66.8& 86.8\\
\hline
\noalign{\smallskip}
\multirow{5}{*}{Clip} & \multirow{2}{*}{Original} &Original&\cellcolor{Gray}37.4& \cellcolor{Gray}62.4& 32.8& 56.7& 3.2& 9.4& \cellcolor{Gray}55.9& \cellcolor{Gray}79.1& 35.0& 59.2& 2.2& 6.9\\
\cline{3-15}
\noalign{\smallskip}
& &Proposed&\cellcolor{Gray}38.1& \cellcolor{Gray}62.7& 34.3& 58.4& 4.9& 14.3& \cellcolor{Gray}58.1& \cellcolor{Gray}81.0& 38.2& 62.6& 2.3& 7.5\\
\cline{2-15}
\noalign{\smallskip}
&\multirow{2}{*}{Proposed} &Original&\cellcolor{Gray}53.6& \cellcolor{Gray}77.5& 42.4& 67.3& 6.7& 17.1& \cellcolor{Gray}60.8& \cellcolor{Gray}82.9& 43.9& 68.7& 2.5& 7.9\\
\cline{3-15}
\noalign{\smallskip}
&&Proposed&\cellcolor{Gray}\textbf{56.4}& \cellcolor{Gray}\textbf{79.5}& 45.7& 70.3& 12.1& 27.7& \cellcolor{Gray}\textbf{62.2}& \cellcolor{Gray}\textbf{83.9}& 49.0& 73.6& 2.6& 8.3\\
\cline{2-15}
\noalign{\smallskip}
&\multicolumn{2}{l}{Real-valued Net} &\cellcolor{Gray}68.0& \cellcolor{Gray}88.1& 67.5& 87.6& 64.2& 85.3& \cellcolor{Gray}69.7& \cellcolor{Gray}89.1& 67.9& 87.8& 57.1& 79.9\\
\hline
\noalign{\smallskip}
\multicolumn{5}{l}{Full-precision original ResNet\cite{resnet}}& 69.3& 89.2&&&&& 73.3 &91.3\\
\hline
\end{tabular}
%\vspace{-0.8cm}
\end{center}
\end{table}
\setlength{\tabcolsep}{1.4pt}

Based on Table \ref{table:ablation study}, we can evaluate each technique's individual contribution and collective contribution of each unique combination of these techniques towards the final accuracy.

1) Comparing the $4^{\textit{th}}-7^{\textit{th}}$ columns with the $8^{\textit{th}}-9^{\textit{th}}$  columns, both the proposed Bi-Real net and the binarized standard ResNet outperform their plain counterparts with a significant margin, which validates the effectiveness of the shortcut and the disadvantage of directly concatenating the 1-bit convolutional layers. As Plain-18 has a thin and deep structure, which has the same weight filters but no shortcut, binarizing it results in very limited network representational capacity in the last convolutional layer, and can thus hardly achieve good accuracy. 

2) Comparing the $4^{\textit{th}}-5^{\textit{th}}$  and $6^{\textit{th}}-7^{\textit{th}}$ columns, the 18-layer Bi-Real net structure improves the accuracy of the binarized standard ResNet-18 by about 18\%. This validates the conjecture that the Bi-Real net structure with more shortcuts further enhances the network capacity compared to the standard ResNet structure. Replacing the 2-conv-layer-per-block structure employed in Res-Net with two 1-conv-layer-per-block structures, adopted by Bi-Real net, could even benefit a real-valued network.

3) All proposed techniques for initialization, weight update, and activation backward improve the accuracy to various extent. For the 18-layer Bi-Real net structure, the improvement from the weight (about 23\%, by comparing the $2^{\textit{nd}}$ and $4^{\textit{th}}$ rows) is greater than the improvement from the activation (about 12\%, by comparing the $2^{\textit{nd}}$ and $3^{\textit{rd}}$ rows) and the improvement from replacing ReLU with Clip for initialization (about 13\%, by comparing the $2^{\textit{nd}}$ and $7^{\textit{th}}$ rows).  These three proposed training mechanisms are orthogonal to each other and can function collaboratively towards enhancing the final accuracy.

4) The proposed training methods can improve the final accuracy for all three networks in comparison with the original training method, which implies that these proposed three training methods are universally suitable for various networks. 

5) The two implemented Bi-Real nets (\textit{i.e.}, the 18-layer and 34-layer structures) together with the proposed training methods achieve approximately 83\% and 89\% of the accuracy level of their corresponding full-precision networks, but with a huge amount of speedup and computation cost saving. 

In brief, the shortcut enhances the network representational capability, and the proposed training methods help the network approach the accuracy upper bound.

As discussed in Sec. \ref{sec:activation_back}, we investigate a higher-order approximation to the derivative of the sign function. Specifically, we carry out an ablation study to answer the question of how much increase in performance can we obtain by using a higher order approximation to the derivative of the sign function. As shown in Table \ref{table:activation-ablation-study}, the gain from changing the second-order approximation of the sign function to the third-order approximation is diminished, to only a 0.06\% increase in top-1 accuracy. Considering the computational overhead, we suggest using the second-order approximation.

\setlength{\tabcolsep}{1pt}
\begin{table}[t]
\begin{center}
\caption{Top-1 and top-5 accuracies (in percentage) comparison between using the different approximation to the derivative of the sign function on 18-layer Bi-Real net.}
%\vspace{-0.3cm}
\label{table:activation-ablation-study}
\begin{tabular}{cccccc}
\hline\noalign{\smallskip}
& & \ \ \ \ \ ApproxSign \ \ \ \ & Third-order ApproxSign  \\
\noalign{\smallskip}
\hline
 &  Top-1 & 56.40\%  & 56.46\% \\
 &  Top-5 & 79.50\%  & 79.74\% \\
\hline
\end{tabular}
%\vspace{-0.8cm}
\end{center}
\end{table}
\setlength{\tabcolsep}{1.4pt}

%\vspace{-0.2cm}
\setlength{\tabcolsep}{1pt}
\begin{table}[t]
\begin{center}
\caption{This table compares both the top-1 and top-5 accuracies of our Bi-Real net with other state-of-the-art binarization methods: BinaryNet \cite{binarynet} , XNOR-Net \cite{xnornet}, and ABC-Net \cite{dji} on both the Res-18 and Res-34 \cite{resnet}. The Bi-Real net outperforms the other methods by a considerable margin.}
%\vspace{-0.3cm}
\label{table:accuracy_comparison}
\begin{tabular}{cccccccc}
\hline\noalign{\smallskip}
& & Bi-Real net & BinaryNet\cite{binarynet} & ABC-Net\cite{dji}  & XNOR-Net\cite{xnornet}  & Full-precision\cite{resnet} \\
\noalign{\smallskip}
\hline
\multirow{2}{*}{18-layer}  & \ Top-1 &  56.4\%  & 42.2\% & 42.7\% & 51.2\% & 69.3\% \\
 & \ Top-5 & 79.5\%   & 67.1\% & 67.6\% & 73.2\% & 89.2\% \\
\hline
\multirow{2}{*}{34-layer} & \ Top-1 & 62.2\%   & -- & 52.4\% & -- &73.3\% \\
& \ Top-5 & 83.9\%  & -- & 76.5\% & -- &91.3\% \\
\hline
\end{tabular}
%\vspace{-0.8cm}
\end{center}
\end{table}
\setlength{\tabcolsep}{1.4pt}

\subsection{Accuracy Comparison with State-of-the-Arts}
%\vspace{-0.1cm}
\label{sec:accuracy_comparison}
While the ablation study demonstrates the effectiveness of our 1-layer-per-block structure and the proposed techniques for optimal training, it is also necessary to make a comparison with other state-of-the-art methods to evaluate Bi-Real net's overall performance. To this end, we carry out a comparative study with three methods: BinaryNet \cite{binarynet}, XNOR-Net \cite{xnornet}, and ABC-Net \cite{dji}. These three networks are representative methods of binarizing both weights and activations for CNNs and achieve state-of-the-art results. Note that for fair comparison, our Bi-Real net contains the same number of weight filters as the corresponding ResNet that these methods attempt to binarize, differing only in the shortcut design.

Table \ref{table:accuracy_comparison} shows the results. The results of the three networks are quoted directly from the corresponding references, except the result of BinaryNet, which is quoted from ABC-Net \cite{dji}. 
The comparison clearly indicates that the proposed Bi-Real net outperforms the three networks by a considerable margin in terms of both the top-1 and top-5 accuracies. Specifically, the 18-layer Bi-Real net outperforms its 18-layer counterparts BinaryNet and ABC-Net with a relative 33\% advantage and achieves a roughly 10\% relative improvement over XNOR-Net. Similar improvements can be observed for the 34-layer Bi-Real net. In short, our Bi-Real net is more competitive than the state-of-the-art binary networks.   
\setlength{\tabcolsep}{1pt}
\begin{table}[t]
\begin{center}
\caption{This table compares both the top-1 and top-5 accuracies of our Bi-Real net with binarization methods in BinaryNet \cite{binarynet} and XNOR-Net~\cite{xnornet} on Res-50 and Res-152 \cite{resnet}. The results of XNOR-Net is quoted from~\cite{label-refinery}. The Bi-Real net outperforms BinaryNet by a considerable margin and achieves comparable results as the XNOR-Net with a simpler binarization function.}
%\vspace{-0.3cm}
\label{table:accuracy-deep-network}
\begin{tabular}{cccccc}
\hline\noalign{\smallskip}
& &Bi-Real net \ \ & BinaryNet\cite{binarynet}\ \ &  XNOR-Net\cite{xnornet} \ \  & Full-precision Res-Net\cite{resnet} \ \ \\
\noalign{\smallskip}
\hline
\multirow{2}{*}{50-layer}  & \ Top-1 & 62.6\% & 9.4\% & 63.1\% & 74.7\% \\
 & \ Top-5  & 83.9\% & 22.4\% & 83.6\% & 92.1\% \\
\hline
\multirow{2}{*}{152-layer} & \ Top-1 & 64.5\% & 8.9\% & -- & 76.5\% \\
& \ Top-5 & 85.5\% & 21.0\% & -- & 93.2\% \\
\hline
\end{tabular}
%\vspace{-0.8cm}
\end{center}
\end{table}
\setlength{\tabcolsep}{1.4pt}

Table \ref{table:accuracy-deep-network} shows the results of Bi-Real net on the deeper network with the bottleneck structure. Our Bi-Real net contains the same number of weight filters as the corresponding ResNet. We re-implement the method proposed in BinaryNet on ResNet-50 and ResNet-152. For fair comparison, both methods use the same data augmentation with the lower dimension of the image randomly sampled in [256,480] while keeping the aspect ratio intact. A random crop of size 224 $\times$ 224 is selected from the rescaled image or its horizontal flip. The results show that Bi-Real net with an ultra-deep network structure outperforms the method in BinaryNet~\cite{binarynet} by a large margin. The results also show that without adding extra shortcuts to preserve every layer's real-valued output, the 1-bit CNNs can hardly scale up to 152-layers deep, while our proposed Bi-Real net can achieve 64.5\% top-1 accuracy. Bi-Real net also achieves comparable accuracy to XNOR-Net~\cite{xnornet} on the 50-layer structures. Compared to XNOR-Net, we do not need to store or compute the real-valued scaling factor for multiplying with the binary weights and activations, which makes our network easier for implementation and execution.

\subsection{Efficiency and Memory Usage Analysis}
\label{sec:efficiency_comparison}
In this section, we analyze the saving of memory usage and speedup in computing Bi-Real net by both theoretical analysis and real-world estimation on FPGA devices.

\subsubsection{Resource computation}

The memory usage is computed as the summation of 32 bits times the number of real-valued parameters and 1 bit times the number of binary parameters in the network. For efficiency comparison, following the suggestion of Uniq~\cite{uniq}, we use BOPs to measure the total multiplication computation in Bi-Real net. BOP refers to the number of bit-operations in a neural network, where the calculation method is the same as calculating FLOPs for the floating-point operations in \cite{resnet}, excepting the operation calculated in BOPs is bit-wise.

\setlength{\tabcolsep}{1pt}
\begin{table}[t]
\begin{center}
\caption{Memory usage and BOPs calculation in Bi-Real net.}
%\vspace{-0.2cm}
\label{table:memoy_n_flop}
\begin{tabular}{ccccccc}
\hline\noalign{\smallskip}
& & Memory usage \ & \ Memory saving &  \ BOPs &  \ Speedup \\
\noalign{\smallskip}
\hline
\multirow{3}{*}{18-layer}  & \ Bi-Real net&  33.6 Mbit & 11.14$\times$ &1.04 $\times 10^{10}$ & 11.06$\times$  \\
 & \ XNOR-Net\cite{xnornet} & 33.7 Mbit & 11.10$\times$ &1.07 $\times 10^{10}$ & 10.86$\times$ \\
 & \ Full-precision Res-Net\cite{resnet} & 374.1 Mbit & -- &1.16 $\times 10^{11}$ & --\\
\hline
\multirow{3}{*}{34-layer}  & \ Bi-Real net & 43.7 Mbit & 15.97$\times$ &124 $\times 10^{10}$ & 18.99$\times$  \\
 & \ XNOR-Net\cite{xnornet} & 43.9 Mbit & 15.88$\times$ &127 $\times 10^{10}$ & 18.47$\times$  \\
 & \ Full-precision Res-Net\cite{resnet} & 697.3 Mbit & -- &234 $\times 10^{11}$ & --\\

\hline
\end{tabular}
%\vspace{-1.0cm}
\end{center}
\end{table}
\setlength{\tabcolsep}{1.4pt}

We follow the suggestion in XNOR-Net \cite{xnornet}, to keep the weights and activations in the first convolution and the last fully-connected layers to be real-valued. We also adopt the same real-valued 1$\times$1 convolution in the Type B shortcut \cite{resnet}, as implemented in XNOR-Net. Note that this 1$\times$1 convolution is for the transition between two stages of ResNet and thus all information should be preserved. As the number of weights in those three kinds of layers accounts for only a very small proportion of the total number of weights, the limited memory saving for binarizing them does not justify the performance degradation caused by the information loss.

For both the 18-layer and the 34-layer networks, the proposed Bi-Real net reduces the memory usage by 11.1 times and 16.0 times, respectively, and achieves computation reduction of about 11.1 times and 19.0 times, in comparison with the full-precision counterparts. 
Without using real-valued weights and activations for scaling binary ones during the inference time, our Bi-Real net requires fewer BOPs and uses less memory than XNOR-Net and is also much easier to implement.

\subsubsection{On-board speed estimation}
We estimate the execution time of an 18-layer Bi-Real net as well as its real-valued counterpart on FPGA (Field-Programmable Gate Array) with the Vivado Design Suite~\cite{vivado}, which is targeted at embedded applications. Table~\ref{table:fpga} shows the speed and resource usage comparison of the entire networks. The proposed Bi-Real net achieves a 6.07$\times$ speed up with the same or fewer resources compared to its real-valued counterpart on an FPGA board. Table~\ref{table:FPGA_board} shows the speed estimation of each individual module. Binary convolutional layers achieve a 15.8$\times$ speed up compared with real-valued convolutions. By adding up the execution time of all the operations, 18-layer Bi-Real net is able to achieve a 7.38$\times$ speedup than the real-valued network with the same structure.

\setlength{\tabcolsep}{1pt}
\begin{table}[t]
\begin{center}
\caption{Execution time and resource usage comparison between inferring the 18-layer Bi-Real net and its real-valued counterpart on FPGA. The codes and other implementation details will be released at https://github.com/liuzechun/Bi-Real-net.}
\label{table:FPGA_board}
\begin{tabular}{cccccc}
\noalign{\smallskip}
\hline\noalign{\smallskip}
& Execution Time(ms)  & \multicolumn{3}{c}{Resource usage}\ \ \  \\
\noalign{\smallskip}
&  & \ \ \ LUT \ \ \ & \ \ \ FF \ \ \ & \ \ BRAM \ \ \\
\noalign{\smallskip}
\hline\noalign{\smallskip}
32-bit network & 2394.6 & 256207 & 315644 & 2086 \\
1-bit network & 394.7 & 154036 & 197606 & 2086 \\
\noalign{\smallskip}
\hline\noalign{\smallskip}
Time/Resource reduction ratio & 6.07$\times$ & 1.66$\times$ & 1.60$\times$ & 1$\times$ \\
\noalign{\smallskip}
\hline
\end{tabular}
\end{center}
\end{table}
\setlength{\tabcolsep}{1.4pt}

\setlength{\tabcolsep}{1pt}
\begin{table}[t]
\begin{center}
\caption{Speed comparison of individual modules in 18-layer Bi-Real net with 1-bit convolutions and 32-bit convolutions on FPGA.}
\label{table:fpga}
\begin{tabular}{cccccc}
\hline\noalign{\smallskip}
&  \multicolumn{2}{c}{Execution time} & Speedup ratio \ \ \  \\
\noalign{\smallskip}
& 1-bit network & \ \ \ \ 32-bit network \ \ \ \ & \\
\noalign{\smallskip}
\hline
3 $\times$ 3  Convolutional layers & 2.55s  & 24.23s & 15.8$\times$ \\
Downsampling layers & 0.13s & 0.13s & -- \\
First convolutional layer & 0.08s & 0.08s & --\\
Fully-connected layer & 1.02ms & 1.02ms & --\\
BatchNorm layers & 0.62s & 0.62s & -- \\
All operations considered & 3.4s &  25.1s & 7.38$\times$ \\
\hline
\end{tabular}
\end{center}
\end{table}
\setlength{\tabcolsep}{1.4pt}

\subsection{Application: Pixelwise Depth Estimation}
\label{sec:depth_estimation}
Depth estimation is an important task for autonomous driving and drone navigation. Compressing a depth estimation CNN is crucial to deploying the powerful CNN to mobile devices which have limited memory and computational resources. In this section, we replace the real-valued Res-50 network in \cite{unsupervised-cnn-depth} with a 50-layer Bi-Real net for pixelwise depth estimation.

The experimental evaluations were carried out on the KITTI dataset \cite{kitti}, which contains pictures of the roads captured using a stereo camera mounted on a moving vehicle. We employed the same training/validation split and data pre-processing method as \cite{unsupervised-cnn-depth} for fair comparison.

In the training phase, the images were resized to 160 $\times$ 608 and no data augmentation was applied. We trained the network for 60K iterations with a mini-batch size of 10. We started from a learning rate of 0.001 and divided it by 10 at every 30K iterations. We pre-trained the real-valued network and then used it to initialize and fine-tune the binarized network.

In the testing phase, we evaluated our results on the same cropped region of interests as \cite{unsupervised-cnn-depth} and compared the depth prediction results with the corresponding ground-truth depth maps.

\setlength{\tabcolsep}{1pt}
\begin{table}[t]
\begin{center}
\caption{Comparison on accuracy of the 50-layer Bi-Real net and binarizing full-precision network~\cite{unsupervised-cnn-depth} with the method proposed in BinaryNet~\cite{binarynet}. The accuracy of the network with full-precision resolution~\cite{unsupervised-cnn-depth} is also included for reference.}
\label{table:depth-estimation}
\begin{tabular}{cccccc}
\hline\noalign{\smallskip}
\multicolumn{3}{c}{Depth Estimation Network} \\
\noalign{\smallskip}
\hline
\noalign{\smallskip}
 Bi-Real net\ \ \ \  &  BinaryNet \cite{binarynet} \ \ \ \ & Full-precision network \cite{unsupervised-cnn-depth} \\
\noalign{\smallskip}
\hline
\noalign{\smallskip}
 84.9\% & 83.0\% & 85.2\% \\
\noalign{\smallskip}
\hline
\end{tabular}
\vspace{-0.5cm}
\end{center}
\end{table}
\setlength{\tabcolsep}{1.4pt}
The results show that Bi-Real net achieves comparable accuracy to the real-valued network proposed in \cite{unsupervised-cnn-depth} and is 2\% higher than directly binarizing the original network with the method in \cite{binarynet}. The results provide a piece of evidence that the proposed Bi-Real net not only works well on classification tasks but can also be applied to other regression tasks like pixelwise depth estimation.

\section{Conclusions}
In this study, we proposed a novel 1-bit CNN model, dubbed Bi-Real net. Compared to standard 1-bit CNNs, Bi-Real net utilizes a simple yet effective shortcut to significantly enhance the representational capability of the 1-bit CNNs. Furthermore, an advanced training algorithm was designed for training 1-bit CNNs (including Bi-Real net), including a tighter approximation to the derivative of the sign function with respect to the activation, a magnitude-aware binarization with respect to the weight, as well as a novel initialization and a two-step training algorithm for deep 1-bit CNNs. 
The extensive experimental results demonstrate that the proposed Bi-Real net and novel training algorithm achieve superior results over the state-of-the-art methods and are viable for real-world applications. 

\section{Acknowledgements}
The authors would like to acknowledge HKSAR RGC’s funding support under grant GRF-16203918.
We also would like to thank Zhuoyi Bai, Tian Xia, Prof. Zhenyan Wang and Xiaofeng Hu from Huazhong University of Science and Technology for their efforts in implementing Bi-Real net on FPGA and carrying out the on-board speed estimation.

\bibliographystyle{splncs04}
\bibliography{egbib}
\end{document}